\documentclass[letterpaper, 10 pt, conference]{ieeeconf}  % Comment this line out if you need a4paper
\IEEEoverridecommandlockouts                              % This command is only needed if 
\usepackage{url} % URLs
\usepackage{tabularx,longtable,multirow}%hangcaption
\usepackage{mathptmx} % assumes new font selection scheme installed
\usepackage{amsfonts}
\usepackage{color}
\usepackage{amsmath} % assumes amsmath package installed
\usepackage{amssymb}  % assumes amsmath package installed
\usepackage[utf8]{inputenc}
\usepackage[english]{babel}
\definecolor{qb}{RGB}{128,0,0}
\usepackage{graphicx}
\usepackage{graphics}
\usepackage[noend]{algpseudocode}
\usepackage[ruled,vlined]{algorithm2e}
\usepackage[]{hyperref}
\usepackage{url} % URLs
\usepackage[capitalise,nameinlink]{cleveref}
\include{pythonlisting}

\newtheorem{definition}{Definition}

\title{\LARGE \bf
3D Path Planning from a Single 2D Fluoroscopic Image for Robot Assisted Fenestrated Endovascular Aortic Repair}

\author{Jian-Qing Zheng$^{1}$, Xiao-Yun Zhou$^{1}$, Celia Riga$^{2}$ and Guang-Zhong Yang$^{1}$% <-this % stops a space
\thanks{$^{1}$Jian-Qing Zheng, Xiao-Yun Zhou, and Guang-Zhong Yang are with the Hamlyn Centre for Robotic Suegery, Imperial College London, UK.
        {\tt\small j.zheng17@imperial.ac.uk}}%
\thanks{$^{2}$Celia Riga is with the Regional Vascular Unit, St Mary's Hospital, London, UK and the Academic Division of Surgery, Imperial College London, UK}%
}

\begin{document}

\maketitle
\thispagestyle{empty}
\pagestyle{empty}

\begin{abstract}
% Currently, 2D fluoroscopic images are used for navigation in Fenestrated Endovascular Aortic Repair (FEVAR) where 3D alignment is required. The navigation with insufficient anatomical information from 2D images indicates both longer operation duration and higher radiation dose. In this paper, 3D robotic path planning is achieved in real-time based on only a single 2D fluoroscopic image of Abdominal Aortic Aneurysm (AAA). Graph matching is proposed for the correspondence between the pre-operative 3D and intra-operative 2D AAA skeletons. Then skeleton deformation with regularization in terms of skeleton length and smoothness was used to register the pre-operative 3D AAA skeleton to intra-operative 2D AAA skeleton. Furthermore, deep learning was applied to segment pre-operative 3D AAA from Computed Tomography (CT) scans to facilitate the framework automation. Simulation, phantom and patient AAA data were used to validate the proposed framework. 3D distance error of 2mm was achieved in the phantom setup. Performance advantages were also achieved in terms of accuracy, robustness and time-efficiency. All the codes will be open-source.
The current standard of intra-operative navigation during Fenestrated Endovascular Aortic Repair (FEVAR) calls for need of 3D alignments between inserted devices and aortic branches. The navigation commonly via 2D fluoroscopic images, lacks anatomical information, resulting in longer operation hours and radiation exposure. In this paper, a framework for real-time 3D robotic path planning from a single 2D fluoroscopic image of Abdominal Aortic Aneurysm (AAA) is introduced. A graph matching method is proposed to establish the correspondence between the 3D preoperative and 2D intra-operative AAA skeletons, and then the two skeletons are registered by skeleton deformation and regularization in respect to skeleton length and smoothness. Furthermore, deep learning was used to segment 3D pre-operative AAA from Computed Tomography (CT) scans to facilitate the framework automation. Simulation, phantom and patient AAA data sets have been used to validate the proposed framework. 3D distance error of 2mm was achieved in the phantom setup. Performance advantages were also achieved in terms of accuracy, robustness and time-efficiency. All the code will be open source.
\end{abstract}

\section{INTRODUCTION}
\begin{figure}[thpb]
      \centering
      \framebox{\parbox{1.5in}{\includegraphics[width=1.5in]{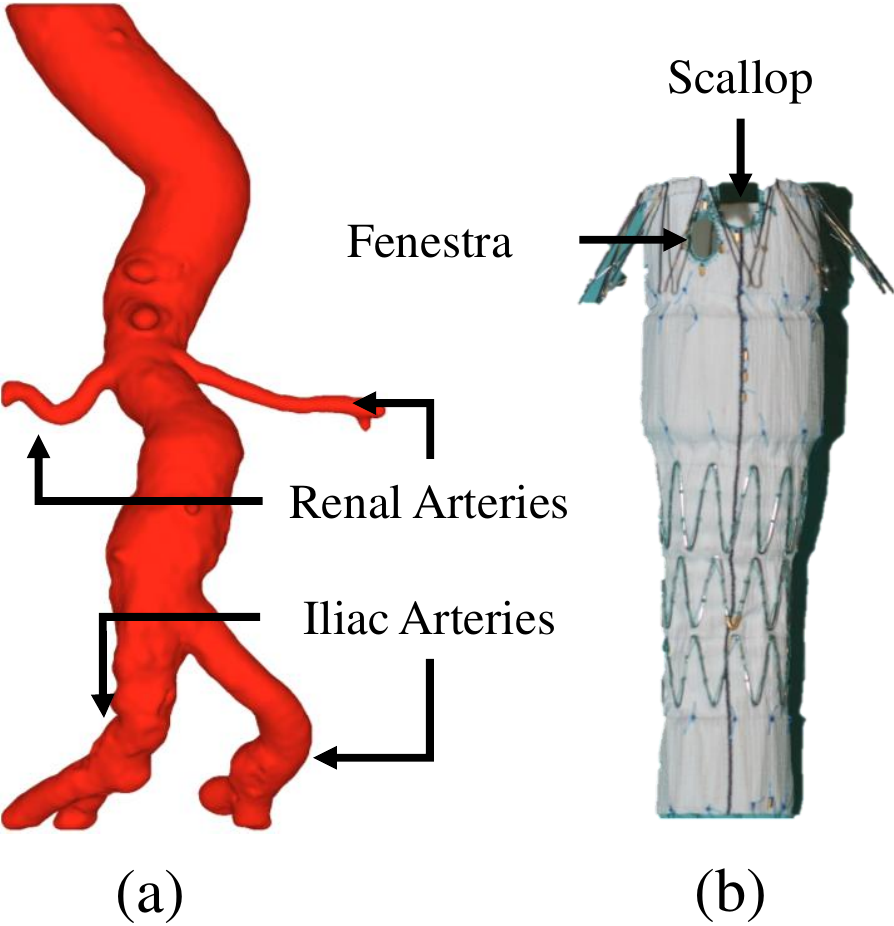}}}
      \caption{An AAA example (a) where the inlarged part is near renal arteries and a fenestrated stent graft example (b) with fenestrations and scallops.}
      \label{fig:introduction}
\end{figure}

Abdominal aortic aneurysm (AAA), where the abdominal aortic diameter is 50\% over the normal diameter, is increasingly common among older people \cite{sakalihasan2005abdominal}. It is usually asymptomatic, but causes 85--90\% mortality rate once ruptured \cite{kent2014abdominal}. Fenestrated Endovascular Aneurysm Repair (FEVAR) is a common treatment for AAA. An example of an AAA and a fenestrated stent graft are shown in \cref{fig:introduction}(a) and \cref{fig:introduction}(b) respectively. The accurate 3D alignment of fenestras or scallops to the aortic branches (i.e. renal arteries,), is considered to be a technical challenge during FEVAR. Despite the development of robotic systems like Magellan robot (Hansen Medical, CA, USA), the navigation is still based on interpretation from 2D fluoroscopic images. The current approach suffers from lack of understanding of 3D anatomical environment at the scene, and therefore involving long operation hours and high radiation exposure. This often leads to interactions between catheters and vessel linings, causing further complications, such as embolization and thrombosis \cite{macdonald2009towards}. Therefore, real-time 3D robotic path planning is essential for improving FEVAR outcomes.

Previously, the 3D shape of fenestrated stent graft was instantiated semi-automatically from a single fluoroscopiy with customized markers, Robust Perspective-n-Point (RPnP) method, and gap interpolation \cite{zhou2018real2}, while later Equally-weighted Focal Unet was used to automate this instantiation process \cite{Zhou2017towards}. It supplied 3D spatial information for the fenestration or scallop, while that for aortic branch are also required for navigation.

Extensive research has been directed to real-time intra-operative 3D AAA non-rigid registration and deformation recovering. A real-time general framework was proposed to instantiate intra-operative 3D anatomical shapes with one 2D image input, based on 3D Statistical Shape Model (SSM), Sparse Principal Component Analysis (SPCA), and Kernel Partial Least Squares Regression (KPLSR) \cite{zhou2018real}. A skeleton-based as-rigid-as-possible method was proposed to adapt the 3D pre-operative AAA shape to intra-operative locations of inserted devices and landmarks, from two fluoroscopic images \cite{toth2015adaption}. A deformable 2D/3D registration method using Thin Plate Spline (TPS) with landmarks was described in \cite{raheem2010non}. A 2D/3D vessel registration method with Iterative Closest Point (ICP), TPS, smoothness and length regularization was proposed with a single fluoroscopy input \cite{groher2009deformable}. A method to recover 3D AAA deformations from two fluoroscopic images used neighbouring smoothness term, length preserving, and 2D distance maps as the constraints \cite{liao2010efficient}.

In this paper, a framework is proposed for achieving real-time 3D robotic path planning for FEVAR with only one 2D fluoroscopic image as the input. Graph matching is used for the correspondence between the 3D pre-operative and 2D intra-operative AAA skeletons. Then non-rigid registration between the matched 2D and 3D skeletons is achieved by 3D skeleton deformation using TPS with length and smoothness as the regularization. To establish a better framework automation, a deep learning method is used for AAA segmentation from preoperative Computer Tomography (CT) scans. Three datasets, simulation, phantom and patient have been used for the validation. The 3D distance error of 2mm has been achieved on phantom data with robustness and time-efficiency.
Details of skeleton extraction, graph matching and skeleton deformation are stated in Sec.\ref{sec:extraction}, Sec.\ref{sec:graph}, and Sec.\ref{sec:nonrigid} respectively. Data collection and experiments are illustrated in Sec.\ref{sec:validation}. Results of the simulation, phantom and patient experiments are shown in Sec.\ref{sec:result}. Discussion and conclusion are in Sec.\ref{sec:discussion} and Sec.\ref{sec:conclusion} respectively.

\section{Methodology}

\begin{figure}[thpb]
      \centering
      \framebox{\parbox{3.3in}{\includegraphics[width=3.3in]{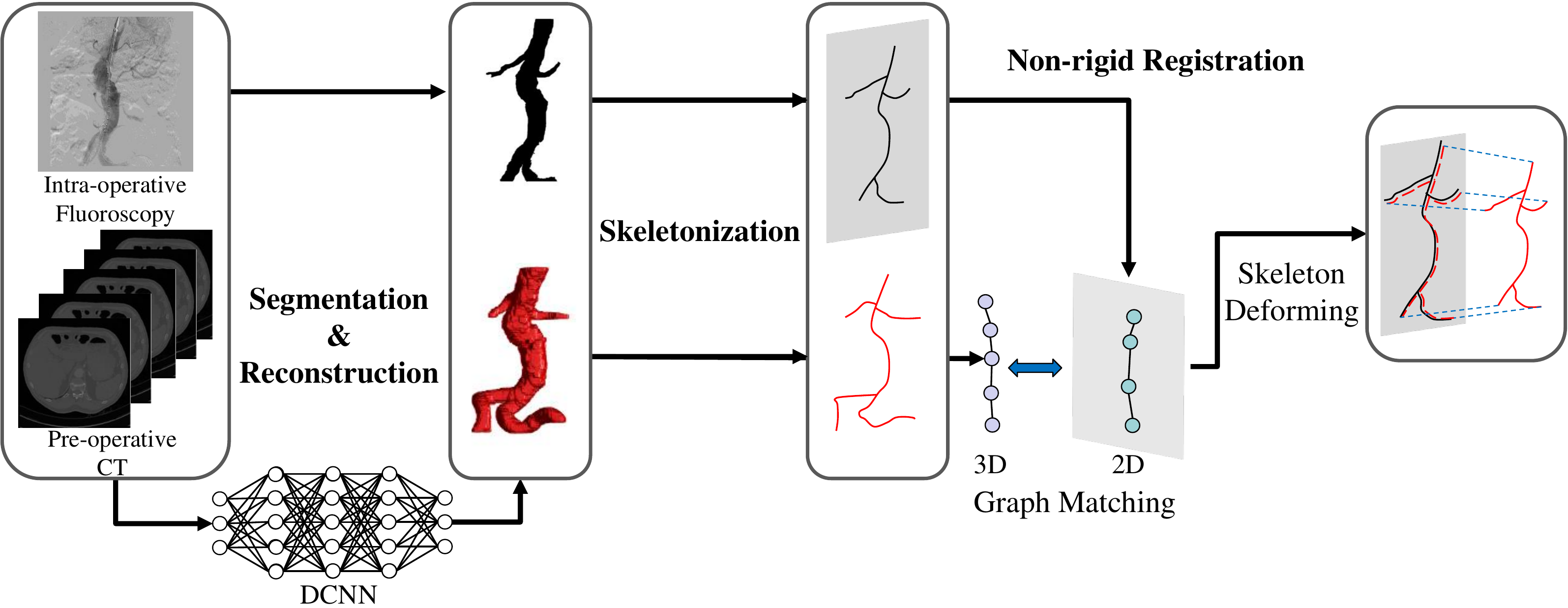}}}
      \caption{Pipeline for the proposed real-time 3D robotic path planning.}
      \label{fig:pipeline}
\end{figure}

A detailed pipeline of the proposed framework for 3D robotic path planning is shown in \cref{fig:pipeline}. Firstly, pre-operative 3D and intra-operative 2D skeletons are extracted from pre-opearative 3D CT scans and intra-operative 2D fluoroscopic images. Then the two skeletons are corresponded with the proposed graph matching. At last, the pre-operative 3D skeleton is deformed with smoothness and length regularization and TPS to match the intra-operative 2D skeleton.

\subsection{Pre-operative 3D and Intra-operative 2D Skeleton Extraction}
\label{sec:extraction}

To improve the framework automation, a 6-block Unet \cite{ronneberger2015u} with convolutional layers, max-pooling layers, and deconvolutional layers was trained to segment the AAA from pre-operative CT scans slice-by-slice. Cross-entropy was computed as the loss function. Stochastic gradient descent was employed for the optimization, with batch size $1$ and momentum $0.9$. The initial learning rate was $0.001$. The loss converged after about $110$k iterations (8$\sim$12 hours training). Linear mapping transformation and window slicing were applied to augment the inter-subject variation in terms of image contrast and AAA position. ${MATLAB}^{\textregistered}$ functions \textit{regionprops} and \textit{bwareopen} were used to automatically extract the largest volume from the segmentation results.

The trained model is used to segment the main aortic aneurysm, while renal and illiac arteries is segmented manually. 3D pre-operative AAA shape is reconstructed from segmented slices with marching cube. 2D intra-operative AAA is segmented manually. The pre-operative 3D AAA and intra-operative 2D AAA were skeletonized with a parallel thinning algorithm for medical axis \cite{kerschnitzki2013architecture}.

\subsection{Graph Matching}
\label{sec:graph}

Pre-operative 3D skeleton points $\textit{\textbf{Y}}=(\textit{\textbf{y}}_1~\cdots~\textit{\textbf{y}}_{\textit{n}})\in\mathbb{R}^{{3}\times{\textit{n}^3}}$ and intra-operative 2D skeleton points $\textit{\textbf{X}}=\begin{pmatrix}\textit{\textbf{x}}_1&\cdots&\textit{\textbf{x}}_{\textit{n}^\textit{2}}\end{pmatrix}\in\mathbb{R}^{\textit{2}\times{\textit{n}^\textit{2}}}$ are registered rigidly $f:\mathbb{R}^{3\times{n}}\to\mathbb{R}^{2\times{n}}$ for a pre-alignment first, with a projection matrix $\textit{\textbf{P}}=\begin{pmatrix} \textit{\textbf{p}}_1	&	\textit{\textbf{p}}_2	&	\textit{\textbf{p}}_3 \end{pmatrix}^\top\in\mathbb{R}^{3\times4}$:
\begin{equation}
\label{eq:projection_points}
f(\textit{\textbf{Y}})=\begin{pmatrix}
    \textit{\textbf{p}}_1^\top\textit{\textbf{Y}}^{\rm h}\oslash\textit{\textbf{p}}_3^\top\textit{\textbf{Y}}^{\rm h}	\\	\textit{\textbf{p}}_2^\top\textit{\textbf{Y}}^{\rm h}\oslash\textit{\textbf{p}}_3^\top\textit{\textbf{Y}}^{\rm h}
	                   \end{pmatrix}
\end{equation}
where $\oslash$ is Hadamard division, and $\textit{\textbf{Y}}^{\rm h}=(\textit{\textbf{y}}_1^{\rm h}~\cdots~\textit{\textbf{y}}_{\textit{n}^3}^{\rm h})=(\textit{\textbf{Y}}^\top~(\textbf{1})_{\textit{n}^3\times1})^\top\in\mathbb{R}^{{4}\times{\textit{n}^3}}$ is the homogeneous vector form of the 3D points. The implementation of this part was introduced in Groher et al' work \cite{groher2009deformable}.

\subsubsection{Skeleton Graph Construction}
We suppose that the 2D/3D skeleton is modeled as a single-component simple graph, $\textit{\textbf{G}}^\textit{d}=\langle\textit{\textbf{V}}^\textit{d},\textit{\textbf{E}}^\textit{d}\rangle, \textit{d}\in\{2,3\}$ with a set of $|\textit{\textbf{V}}^\textit{d}|=\textit{n}^\textit{d}$ nodes $\textit{\textbf{V}}^\textit{d}=\{{v}_\textit{i}^\textit{d}\}_{i\in\textit{\textbf{I}}^d}\subset\mathbb{R}^\textit{d}$ corresponding to $\textit{\textbf{X}}$  or $\textbf{\textit{Y}}$, and the edge set $\textit{\textbf{E}}^\textit{d}\subset\textit{\textbf{V}}^\textit{d}\times\textit{\textbf{V}}^\textit{d}$, in which $\textit{\textbf{I}}^d=[1,n^d]\cap\mathbb{Z}$ is the index set of $\textit{\textbf{V}}^d$. The dimension number $d$ is omitted when an expression is applied with both $d=2$ and $d=3$.

\begin{definition}
Two nodes in the same skeleton graph are adjacent if their Chebyshev distance is less or equal to {\rm 1}:
\begin{equation}
    \textit{{a}}_{\textit{i,j}}:=\left\{
        \begin{array}{ll}
        0, & {\|\textit{\textbf{x}}_\textit{i}-\textit{\textbf{x}}_\textit{j}\|}_{\infty}>0\\
        1, & {\|\textit{\textbf{x}}_\textit{i}-\textit{\textbf{x}}_\textit{j}\|}_{\infty}\leq0
        \end{array}
        \right.
\end{equation}
where $\textit{{a}}_{\textit{i,j}}$ is the $i^{\rm th}$ row $j^{\rm th}$ column element of $\textit{\textbf{G}}$'s adjacency matrix $\textit{\textbf{A}}\in\mathbb{Z}^{{n}\times{n}}$.
\end{definition}

\begin{definition}
The nodes having one adjacent point are defined as end-nodes, with the index number set $\textit{\textbf{I}}_{\rm e}\subseteq\textit{\textbf{I}}$.
The nodes having three or more adjacent nodes are defined as quasi-junction-nodes, with the index number set $\textit{\textbf{I}}_{\rm q-c}\subseteq\textit{\textbf{I}}$:
\end{definition}
    
Construct the subgraph $\textit{\textbf{G}}_{\rm q-c}\subseteq\textit{\textbf{G}}$, including all the junction-nodes and their interconnection edges.
    
\begin{definition}
The node having the largest degree (${\rm deg}(\cdot)$) among each connected component of $\textit{\textbf{G}}_{\rm q-c}$ is defined as a junction-node, with the index number set denoted as $\textit{\textbf{I}}_{\rm c}$. Those nodes which are neither end-nodes nor junction-nodes are defined as connection-nodes.
\end{definition}

\begin{definition}
A sequence of nodes on the shortest path between and including an end-node and the neighbouring junction-node is defined as a branch with the vector of their indexes denoted as $\omega\in\Omega$, in which $\Omega$ is the set of the index number vectors of all the branches. A sequence of nodes on the shortest path between and including two neighbouring junction-nodes is defined as a trunk with the vector of their index numbers denoted as $\pi\in\Pi$, $\Pi$ is the set of the index number vectors of all the trunks.
\end{definition}

The skeleton node sequence of the left/right renal arteries, the left/right illiac arteries and the upper aorta are specified as the five expected preserved branches for registration, which are assumed as the longest five segmented branches. The branches without corresponding arteries or aortas are assumed as the shortest branches and deleted in Alg.~\ref{algo:node_classification} The other branches are omitted in graph matching.

\subsubsection{{Node Classification}}
    \begin{algorithm}[!t]
        \caption{Skeleton Node Classification}
        \label{algo:node_classification}
        \KwIn{$\textit{\textbf{G}}=\langle\textit{\textbf{V}},\textit{\textbf{E}}\rangle$}%$\textit{\textbf{G}}=\langle\textit{\textbf{V}},\o\rangle$}
        \KwOut{$\Omega,\Pi,\textit{\textbf{I}}_{\rm e},\textit{\textbf{I}}_{\rm c},\textit{\textbf{I}}_{\rm d}$}
        \vspace{-0.8em}
        \hrulefill\\
        \nl classify each node as an end- or a quasi-junction-node $\textit{\textbf{I}}_{\rm e}\leftarrow\{{i}|{\rm deg}({v_i})=1\}$,
        $\textit{\textbf{I}}_{\rm q-c}\leftarrow\{{i}|{\rm deg}({v_i})\geq3\}$,$v\in \textit{\textbf{V}}(\textit{\textbf{G}})$\;

        \nl	construct $\textit{\textbf{G}}_{\rm q-c}\subseteq\textit{\textbf{G}}: \textit{\textbf{V}}(\textit{\textbf{G}}_{\rm q-c})\leftarrow\{v_i\}_{i\in\textit{\textbf{I}}_{\rm q-c}},\textit{\textbf{E}}(\textit{\textbf{G}}_{\rm q-c})\leftarrow\{(v_i,v_j)|{i,j\in\textit{\textbf{I}}_{\rm q-c}},i\neq{j}\}\cap\textit{\textbf{E}}(\textit{\textbf{G}})$\;
        \nl \ForEach{\rm connected component of $\textit{\textbf{G}}_{\rm q-c}$ denoted as $C(\textit{\textbf{G}}_{\rm q-c})$}{
        	\nl	$\textit{\textbf{I}}_{\rm c}\leftarrow\textit{\textbf{I}}_{\rm c}\cup\operatorname*{arg\,max}_i\{{\rm deg}(v_i):v_i\in\textit{\textbf{V}}(C(\textit{\textbf{G}}_{\rm q-c}))\}$\;}
        \nl copy $\textit{\textbf{I}}'_{\rm e}\leftarrow{{\textit{\textbf{I}}}_{\rm e}}$ to mark all the end-nodes\;
        \nl initialize $s\leftarrow{0}$, $\Omega\leftarrow{\o}$ and $\textit{\textbf{I}}_{\rm d}\leftarrow{\o}$\;
        \nl \While{$\textit{\textbf{I}}'_{\rm e}\neq\o$}{
            \nl \ForEach{{\rm marked end-node} ${v_i}:{i\in\textit{\textbf{I}}}'_{\rm e}$}{
                \nl \ForEach{{\rm node} $v_j:j\in\textit{\textbf{I}},~g(v_i,v_j)=s$}{ 
                    \nl \If{$j\in\textit{\textbf{I}}_{\rm c}$}{
                        \nl unmark the end-node $\textit{\textbf{I}}'_{\rm e}\leftarrow\textit{\textbf{I}}'_{\rm e}\backslash\{i\}$\;
                        \nl initialize a vector $\omega$\;
                        \nl store the indexes of the shortest path nodes from $v_j$ to $v_i$ into $\omega$\;
                        \nl \eIf{$|\textit{\textbf{I}}'_{\rm e}|>\tau$}{
                            \nl $\textit{\textbf{I}}_{\rm c}\leftarrow\textit{\textbf{I}}_{\rm c}\backslash\{j\}$\;
                            \nl \lIf{$|\omega|<\iota$}{
                                $\textit{\textbf{I}}_{\rm d}\leftarrow\textit{\textbf{I}}_{\rm d}\cup\omega$}}{
                            \nl $\Omega\leftarrow{\Omega\cup\{\omega\}}$\;
                        }\vspace{-0.3em}
                    }\vspace{-0.3em}
                }\vspace{-0.3em}
            }
            \nl $s\leftarrow{s+1}$\;
        }
        \nl copy $\textit{\textbf{I}}'_{\rm c}\leftarrow{{\textit{\textbf{I}}}_{\rm c}}$ to mark all the junction-nodes\;
        \nl initialize $s\leftarrow{0}$, $\Pi\leftarrow{\o}$, and $\sigma_k\leftarrow{\o},~\forall{k\in\textit{\textbf{I}}'_{\rm c}}$\;
        \nl \While{$\textit{\textbf{I}}'_{\rm c}\neq\o$}{
            \nl \ForEach{{\rm junction-node} ${v_i}:{i\in\textit{\textbf{I}}}_{\rm c}$}{
                \nl \ForEach{{\rm node} $v_j:j\in\textit{\textbf{I}}\backslash{},g(v_i,v_j)=s$}{
                    \nl \If{$j\in\sigma_k,~\exists{k\in\textit{\textbf{I}}'_{\rm c}}$}{
                        \nl unmark the junction-node $\textit{\textbf{I}}'_{\rm c}\leftarrow\textit{\textbf{I}}'_{\rm c}\backslash\{k\}$\;
                        \nl initialize a vector $\pi$\;
                        \nl store the indexes of the shortest path nodes from $v_k$ to $v_i$ into $\pi$\;
                        \nl $\Pi\leftarrow{\Pi\cup\{\pi\}}$\;
                        }
                    \nl $\sigma_i\leftarrow{\sigma_i\cup\{j\}}$\;
                }\vspace{-0.3em}
            }
            \nl $s\leftarrow{s+1}$\;
        }
    \end{algorithm}
    
An algorithm with a shortest path algorithm embedded is proposed in Alg.\ref{algo:node_classification} to search and classify branch nodes and trunk nodes.

The branch nodes are searched and classified as preserved, omitted or deleted branches in a nested loop (row 5-19), where the shortest path from each end-node to the neighbouring junction-node is searched based on Breadth First Searching (BFS). If the number of marked end-nodes is not greater than the threshold number of preservation $\tau:=5$, the branch nodes on the found path from the junction-node to the end-node will be preserved. Otherwise, the geodesic length of the found path is compared with the threshold geodesic length of deleted branches $\iota$ for deleting or omitting, and the found junction-node is converted to a connection-node. Then the trunk nodes are searched in the other nested loop (row 20--31), where the shortest path between each two neighbouring junction-nodes is searched based on bidirectional BFS and preserved.
    
The path length between nodes $u$ and $v$ in the Alg.~\ref{algo:node_classification} is calculated using the geodesic distance, denoted as $g(u,v)$, which is different from the path length in the Sec.\ref{sec:branch_trunk_assign}. The geodesic distance calculation and the shortest path searching in these algorithms were implemented based on the adjacency matrix product in ${MATLAB}^{\textregistered}$. It might be more efficient implemented using Dijkstra's algorithm \cite{dijkstra1959note} in other programming languages.
    
\subsubsection{{Branch and Trunk Matching}}
 
The trunks and junction-nodes in 2D and 3D are matched first according to the ranked geodesic distance of these trunks. Then, the tangent vector of each branches $\textit{\textbf{T}}=(\textit{\textbf{t}}_1~\cdots~\textit{\textbf{t}}_{|\Omega|})\in\mathbb{R}^{d\times|\Omega|}$ in 2D and 3D is approached respectively using the vector from the starting point with $\eta_{\rm s}$ as the index number in the branch, to the terminating point with another index $\eta_{\rm t}$. The similarity matrix $\textit{\textbf{S}}:=\textit{\textbf{T}}^\top\textit{\textbf{T}}$ of the tangent vectors is calculated to evaluate the direction difference between the tangents of the branches; According to the similarity matrix and the matched corresponding junction-nodes, the five pairs of 2D and 3D branches are matched respectively.

\subsubsection{{Branch Node and Trunk Node Assigning}}\label{sec:branch_trunk_assign}
The assignment matrix $\textit{\textbf{M}}\in\mathbb{R}_+^{n^2\times{n^3}}$ is initialized as:
\begin{equation}
\textit{m}_{(i,j)}=0, \forall i\in[1,n^2]\cap\mathbb{Z},j\in[1,n^3]\cap\mathbb{Z}
\end{equation}
where $m_(i,j)$ is  the $i^{\rm th}$ row $j^{\rm th}$ column element of $\textit{\textbf{M}}$. 

A fuzzy and partial correspondence called softassigning between the 2D and the 3D nodes at the matched branch pair is achieved based on equal linear euclidean path length of a projected 3D node and the assigned 2D node(s):
\begin{equation}
    \label{eq:branch_assigning_matrix}
    \left\{
    \begin{array}{lll}
    m_{(\omega_{i+1}^2,\omega_j^3)}=\frac{\sum_{k=2}^{j}{{\|{f(\textit{\textbf{y}}_{\omega_{k}^3})}-{f(\textit{\textbf{y}}_{\omega_{k-1}^3})}\|}_2}-\sum_{l=2}^{i}{{\|\textit{\textbf{x}}_{\omega_{l}^2}-\textit{\textbf{x}}_{\omega_{l-1}^2}\|}_2}}{{{\|\textit{\textbf{x}}_{\omega_{i+1}^2}-\textit{\textbf{x}}_{\omega_{i}^2}\|}_2}}
    \\
    m_{(\omega_i^2,\omega_j^3)}=1-m_{(\omega_{i+1}^2,\omega_j^3)}\\
    0\leq m_{(\omega_i^2,\omega_j^3)}\leq1
	\end{array}
    \right.
\end{equation}
where $i\in[1,|\omega^2|]\cap\mathbb{Z},j\in[1,|\omega^3|]\cap\mathbb{Z}$, $\omega^2\in\Omega^2$ and $\omega^3\in\Omega^3$ are the index vectors for the matched 2D and 3D branches,  $\omega_i$ is the $i^{\rm th}$ element of $\omega$, $|\omega|$ is the element number of $\omega$, and define that $\sum_{i=2}^1{(\cdot)}:=0$.
    
A similar softassigning between the 2D and the 3D nodes at the matched trunk pair is achieved based on equal proportion of linear euclidean path length of a 3D node and the assigned 2D node(s):
\begin{equation}
    \label{eq:trunk_assigning_matrix}
    \left\{
    \begin{array}{lll}
    m_{(\pi_{i+1}^2,\pi_j^3)}=\frac{\frac{\lambda^2}{\lambda^3}\sum_{k=2}^{j}{{\|{\textit{\textbf{y}}_{\pi_{k}^3}}-{\textit{\textbf{y}}_{\pi_{k-1}^3}}\|}_2}-\sum_{l=2}^{i}{{\|\textit{\textbf{x}}_{\pi_{l}^2}-\textit{\textbf{x}}_{\pi_{l-1}^2}\|}_2}}{{{\|\textit{\textbf{x}}_{\pi_{i+1}^2}-\textit{\textbf{x}}_{\pi_{i}^2}\|}_2}}
    \\
    m_{(\pi_i^2,\pi_j^3)}+m_{(\pi_{i+1}^2,\pi_j^3)}=1\\
    0\leq m_{(\pi_i^2,\pi_j^3)}\leq1
	\end{array}
    \right.
\end{equation}
where, $i\in[1,|\pi^2|]\cap\mathbb{Z},j\in[1,|\pi^3|]\cap\mathbb{Z}$, $\pi^2\in\Pi^2$ and $\pi^3\in\Pi^3$ are the index vectors for the matched 2D and 3D trunks, $\lambda^2=\sum_{l=2}^{|\pi^2|}{{\|\textit{\textbf{x}}_{\pi_{l}^2}-\textit{\textbf{x}}_{\pi_{l-1}^2}\|}_2}$ and $\lambda^3=\sum_{k=2}^{|\pi^3|}{{\|{\textit{\textbf{y}}_{\pi_{k}^3}}-{\textit{\textbf{y}}_{\pi_{k-1}^3}}\|}_2}$.

\subsection{Skeleton Deformation}
\label{sec:nonrigid}

\subsubsection{Iterative Optimization}
The target of deformation is to obtain displacement vectors of the 3D skleton nodes $\Phi=(\varphi_{1}\quad\cdots\quad\varphi_{{n}^{3}})$ to minimize the energy function $\varepsilon$ \cite{groher2009deformable}:
\begin{equation}
    \varepsilon(\Phi)=\textit{S}_{\rm D}(\Phi)+\alpha\textit{S}_{\rm L}(\Phi)+\beta\textit{S}_{\rm S}(\Phi)
\end{equation}
where $\textit{S}_{\rm D}$ is the difference measure term \cite{groher2009deformable}; $\textit{S}_{\rm L}$ is the length preservation term with the coefficient $\alpha$, adopt from \cite{groher2009deformable}; $\textit{S}_{\rm S}$ is the smoothness constraint with the coefficient $\beta$, adopt from \cite{liao2010efficient}.
The gradient of ${\textit{S}_{\rm D}}$ with respect to $\Phi$ is calculated as:
\begin{equation}
    \begin{aligned}
    \nabla\textit{S}_{\rm D}(\Phi)=&\begin{pmatrix}
    \textit{\textbf{d}}_{1}+\textit{\textbf{d}}_{2}&\textit{\textbf{d}}_{3}+\textit{\textbf{d}}_{4}&\textit{\textbf{d}}_{5}+\textit{\textbf{d}}_{6}
    \end{pmatrix}^{\top}\\
    &\oslash
    \Big({\begin{pmatrix}
    \textbf{1}
    \end{pmatrix}}_{3\times1}\big({\textit{\textbf{p}}_3^\top}{\tilde{\textit{\textbf{Y}}}}^{\rm h}\circ{\textit{\textbf{p}}_3^\top}{\tilde{\textit{\textbf{Y}}}}^{\rm h}\big)\Big)
    \end{aligned}
\end{equation}
where $(\textbf{1})_{3\times1}$ is a $3\times1$ matrix consisting of $1$s, the matrix $\textit{\textbf{D}}=\begin{pmatrix} \textit{\textbf{d}}_{1}&\cdots&\textit{\textbf{d}}_{6} \end{pmatrix}^\top\in\mathbb{R}^{6\times{n^3}}$ is defined as:
\begin{equation}
    \textit{\textbf{D}}:=-\frac{2}{n^3}\big(\textit{\textbf{X}}\textit{\textbf{M}}-f(\tilde{\textit{\textbf{Y}}})\big)\circ({\textbf{\textit{J}}'} {{\tilde{\textit{\textbf{Y}}}}^{\rm h}})
\end{equation}
where $\circ$ is Hadamard product, and $\textit{\textbf{J}}\in\mathbb{R}^{6\times4}$ is defined as
\begin{equation}
    \begin{aligned}
    \textit{\textbf{J}}:=
    &{\begin{pmatrix}
    {\textit{p}_{11}}&{\textit{p}_{21}}&{\textit{p}_{12}}&{\textit{p}_{22}}
    &{\textit{p}_{13}}&{\textit{p}_{23}}
    \end{pmatrix}}^\top
    {\textit{\textbf{p}}_3}^\top\\
    &-
    {\begin{pmatrix}
    {\textit{p}_{31}}&{\textit{p}_{32}}&{\textit{p}_{33}}
    \end{pmatrix}}^\top\otimes
    {\begin{pmatrix}
    {\textit{\textbf{p}}_{1}}&{\textit{\textbf{p}}_{2}}
    \end{pmatrix}}^\top
    \end{aligned}
\end{equation}
in which, $\otimes$ is the kronecker product.

Using the function $fmincon$ in the optimization toolbox (ver. 6.1) of ${MATLAB}^{\textregistered}$, multiple optimization algorithms have been compared in this non-linear programming problem, including interior algorithm \cite{waltz2006interior}, sequential quadratic programming (SQP) algorithm \cite{gill2005snopt} and Broyden-Fletcher-Goldfarb-Shanno (BFGS) method \cite{powell1978convergence}. The BFGS achieved the minimal time consumption and thus is applied in the proposed method.
    
\subsubsection{Post-deformation}
After the deformation of the assigned 3D skeleton nodes, the unassigned nodes of the 3D skeleton are deformed using TPS \cite{bookstein1989principal} with the nodes of the connected branches or trunks as the controlling points. This part is implemented by the function $TPS3D$ \cite{tps3d} from the file exchange of ${MATLAB}^{\textregistered}$. 

\begin{algorithm}[htb]
    \caption{Deformable 2D-3D Registration}
    \label{algo:deform_regist}
    \KwIn{$\textit{\textbf{X}}$, $\textit{\textbf{Y}}$, $\textit{\textbf{R}}$, $\textit{\textbf{T}}$}
    \KwOut{$\Phi$, $\textit{\textbf{I}}_{\rm d}$}
    \vspace{-0.8em}
    \hrulefill\\
    \nl	construct $\textit{\textbf{G}}^3=\langle\textit{\textbf{V}}^3,\textit{\textbf{E}}^3\rangle$ for $\textit{\textbf{Y}}$\;
    \nl	$\big(\Omega^3,\Pi^3,\textit{\textbf{I}}_{\rm e}^3,\textit{\textbf{I}}_{\rm c}^3,\textit{\textbf{I}}_{\rm d}^3\big)\leftarrow{\rm NODE\_CLASSIFICATION\big(\textit{\textbf{G}}^3\big)}$\;
    \nl	construct $\textit{\textbf{G}}^2=\langle\textit{\textbf{V}}^2,\textit{\textbf{E}}^2\rangle$ for $\textit{\textbf{X}}$\;
    \nl	$\big(\Omega^2,\Pi^2,\textit{\textbf{I}}_{\rm e}^2,\textit{\textbf{I}}_{\rm c}^2,\textit{\textbf{I}}_{\rm d}^2\big)\leftarrow{\rm NODE\_CLASSIFICATION\big(\textit{\textbf{G}}^2\big)}$\;
    \nl	match branches and trunks between $\textit{\textbf{G}}^2$ and $\textit{\textbf{G}}^3$\;
    \nl	initialize the assignment matrix $\textit{\textbf{M}}
\leftarrow{(\textbf{0})}_{{n^2}\times{n^3}}
            $\;
    \nl \ForEach{{\rm matched} $\omega^2\in\Omega^2$ {\rm and} $\omega^3\in\Omega^3$}{
    \nl compute (\ref{eq:branch_assigning_matrix}) for branch node assigning\;}
    \nl \ForEach{{\rm matched} $\pi^2\in\Pi^2$ {\rm and} $\pi^3\in\Pi^3$}{
    \nl compute (\ref{eq:trunk_assigning_matrix}) for trunk node assigning\;}
    \nl	\If{ $\sum_{i=1}^{n^2}{m_{i,j}}=0$}{
    \nl	label $\textit{\textbf{y}}_j$ as unassigned, $\forall{j}\in\textit{\textbf{I}}^3\backslash{\textit{\textbf{I}}}_{\rm d}^3$\;}
    \nl initialized displacement vectors for assigned nodes, $\Phi'$\;
    \nl	\While{\rm not convergence}{
    \nl	compute {$\nabla\varepsilon(\Phi')$} and update $\Phi'$\;}
    \nl compute the displacement vectors $\Phi$ using branch-wise 3D TPS for the unassigned nodes\;
        % \nl \Return{$\Phi$, $\textit{\textbf{I}}_{\rm d}$.}
\end{algorithm}
An overall algorithm for registering pre-operative 3D and intra-operative 2D skeletons is shown in Alg.~\ref{algo:deform_regist}, including graph matching and skeleton deformation, where the skeleton graph is costructed (row 1,3), graph nodes are searched and classfied (row 2,4), branches and trunks are matched in 2D and 3D (row 5) and the nodes are softassigned (row 6--12), and then the 3D skeleton is deformed (row 13--16).

\subsection{Experiment and Validation}
\label{sec:validation}
\begin{figure}[ht]
     \centering
      \framebox{\parbox{2.9in}{\includegraphics[width = 2.9in]{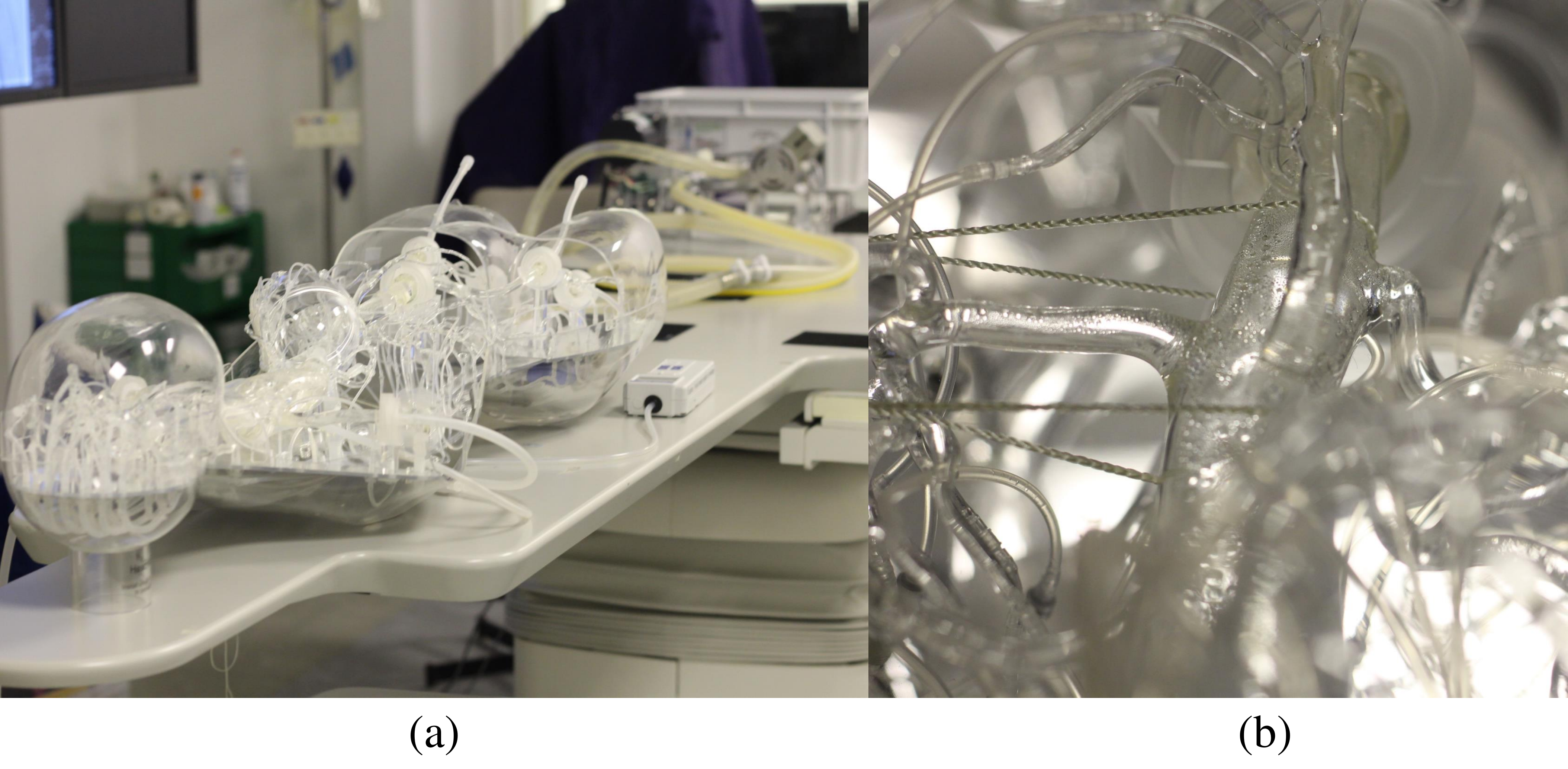}}}
      \caption{(a) Experiments on a phantom, with a pumper simulating the blood circulation; (b) Deformation of a phantom, using a string}
      \label{fig:phantom_deform}
\end{figure}

Five phantom datasets were collected by deforming the endovascular evaluator (${BR~ Biomedicals}^{\textregistered}$ Pvt. Ltd.) using a string, with the experimental setup illustrated in \cref{fig:phantom_deform}. To simulate the blood circulation, 7L water mixed with 500mg contrast dose was pumped in the phantom. CT ($0.453$mm pixel spacing, $1$mm slice thickness, 70mmHg hydraulic pressure) and fluoroscopic ($0.507$mm pixel spacing, 100mmHg hydraulic pressure) images were scanned by GE Innova 4100 (${General Electric}^{\textregistered}$ Healthcare, Bucks, UK). The hydraulic pressure for CT scan was adjusted to be lower for a clear scan. Both 3D CT scans and 2D fluoroscopic images were segmented manually for the phantom data. 
Seven pre-operative AAA patient CT scans ($0.734\sim0.977$mm pixel spacing, $1$mm slice thickness) were collected with Seimens CT VA1 Dummy (${Seimens}^{\textregistered}$, Pic.). Four of them were with corresponding intra-operative fluoroscopic images (pixel spacing is not recorded) and were contrast-enhanced. The AAA of three patients without corresponding intra-operative fluoroscopic images were segmented with \textit{Analyze} (AnalyzeDirect, Inc, Overland Park, KS, USA) and used to train the Unet. The trained model was applied to segmentation on the other four patients with manual corrections at the illiac part.

2D and 3D point-to-curve distances implemented using the function $distance2curve$ from the file exchange of ${MATLAB}^{\textregistered}$ \cite{dist2curve} were used for evaluation. Usually physical distance in mm was calculated for the 2D distance error. Since the fluoroscopic pixel spacing is missed in some cases, distant in pixel was used. 

In our experiments, the $\tau$, $\eta_{\rm s}$, $\eta_{\rm t}$,  $\alpha$  and $\beta$  are set as $5$, $1$, $10$, $500$ and $10$ respectively. The deformable 2D/3D registration approach - ICP-TPS proposed by Grother et al \cite{groher2009deformable} is used as the baseline, where the $\alpha$, $\beta_{\rm final}$, $T_{\rm init}$, $T_{\rm final}$, $T_{\rm update}$ are $0.01$, $3\times10^{-7}$, $500$, $1$ and $0.93$ respectively. %The self-implemented version of the ICP-TPS might not achieve equal performance in their paper, since the code of they work hasn't been open-source yet.

\begin{figure}[ht]
    \centering
    \framebox{\parbox{3.3in}{\includegraphics[width = 3.3in]{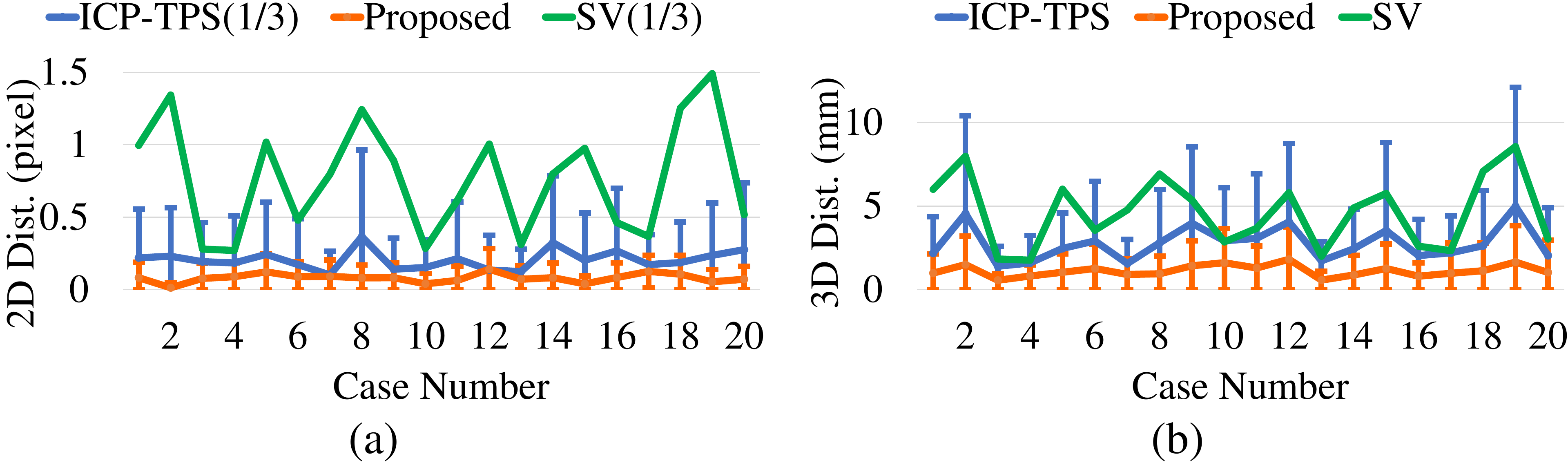}}}
    \caption{The comparison of the performance in 20 simulation cases between the proposed method, ICP-TPS and the shape variation (SV), with mean$\pm$std of 2D(a)/3D(b) distance errors in 2D(a)/3D(b) environment.}
    \label{fig:sim_result}
\end{figure}

\section{Results}
\label{sec:result}

\subsection{Simulation Data Experiment} \label{sec:result_sim}
To purely evaluate the performance of graph matching and skeleton deformation, pre-operative 3D phantom AAA skeletons was projected onto 2D planes to simulate intra-operative 2D AAA skeletons. Each pre-operative 3D phantom AAA skeleton was cross-paired with the other four simulated intra-operative 2D AAA skeletons, totally generating 20 cases. The performance of the proposed method in 2D and 3D environment are presented in \cref{fig:sim_result}(a) and \cref{fig:sim_result}(b) with mean$\pm$std 2D/3D distance errors in 2D/3D environment, which outperformed ICP-TPS in these cases.

\begin{figure}[th]
    \centering
    \framebox{\parbox{3.3in}{\includegraphics[width = 3.3in]{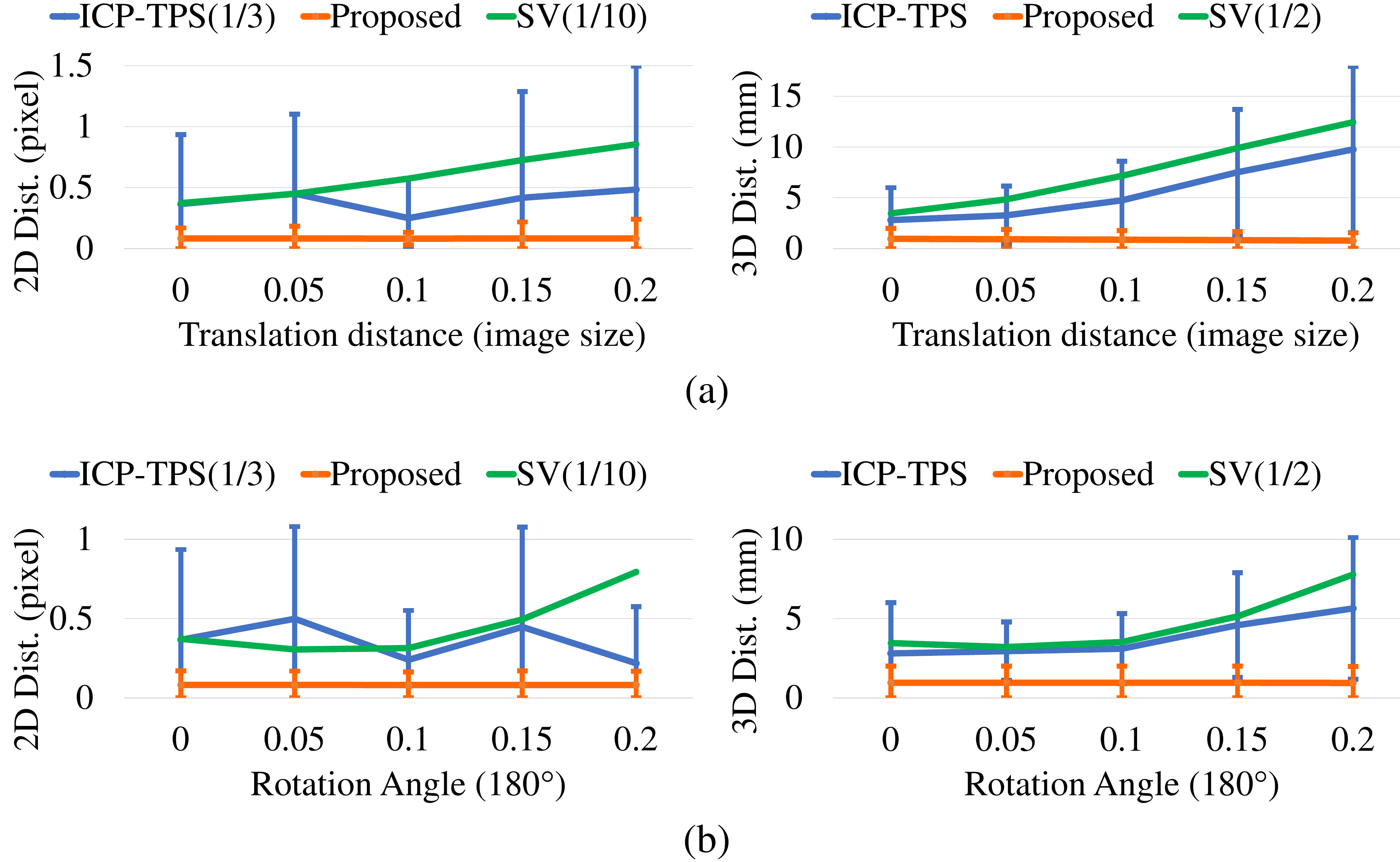}}}
    \caption{The comparison of the performance in one simulation data with additional translations (a) and rotations (b) between the proposed method, ICP-TPS and the shape variation (SV), with mean$\pm$std of 2D/3D distance errors in 2D(left)/3D(right) environment.}
    \label{fig:transf_result}
\end{figure}

\begin{figure}[th]
    \centering
    \framebox{\parbox{2.8in}{\includegraphics[width = 2.8in]{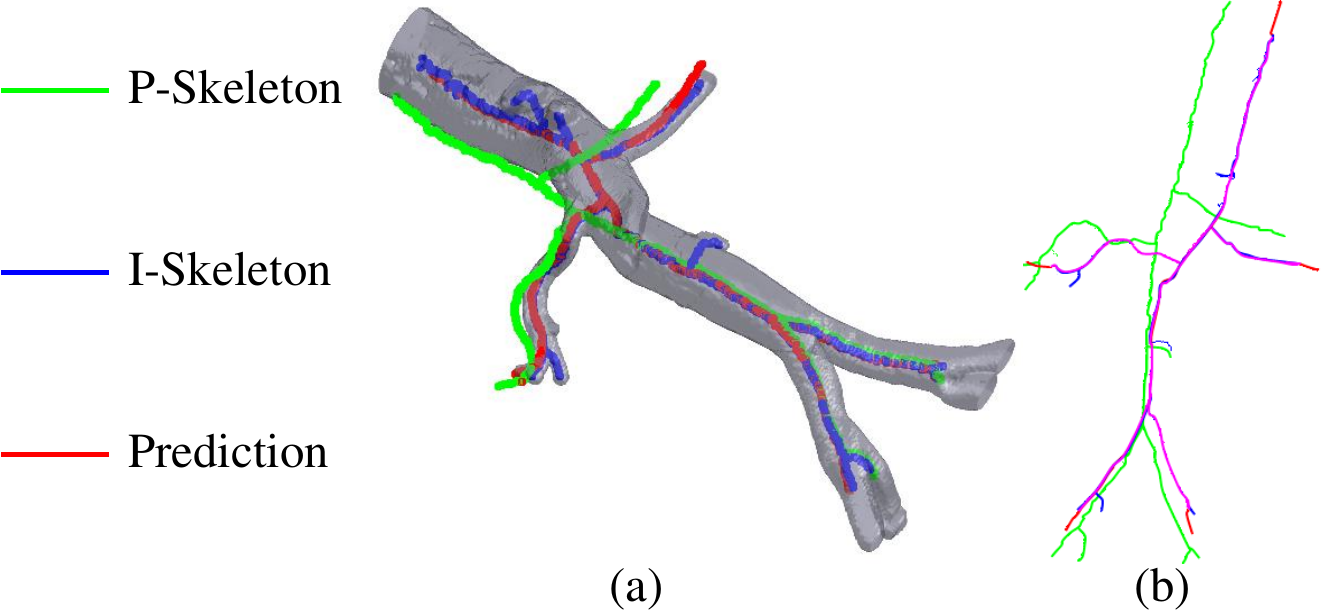}}}
    \caption{The demo of non-rigid registration for the simulation data in 3D (a) and projected 2D (b), where P-Skeleton - pre-operative skeleton, I-Skeleton - intra-operative skeleton, Prediction - the predicted skeleton.}
    \label{fig:sim_example}
\end{figure}

Transformations including translations and rotations in the frontal plane, were added to the simulated intra-operative 2D AAA skeletons to validate the robustness of the proposed method. The performance of proposed method in these cases are presented in \cref{fig:transf_result} with mean$\pm$std 2D and 3D distance errors in 2D and 3D environment. The results show that the proposed method achieved higher robustness with lower 2d/3D distance errors, in theses cases than those by ICP-TPS.
The \cref{fig:sim_example} illustrates a demo of the non-rigid registration by the proposed method in the experiment based on simulation data , where proper registration was achieved with predicted deformation.
\subsection{Phantom Data Experiment} \label{sec:result_phan}
\begin{figure}[th]
    \centering
    \framebox{\parbox{3.3in}{\includegraphics[width = 3.3in]{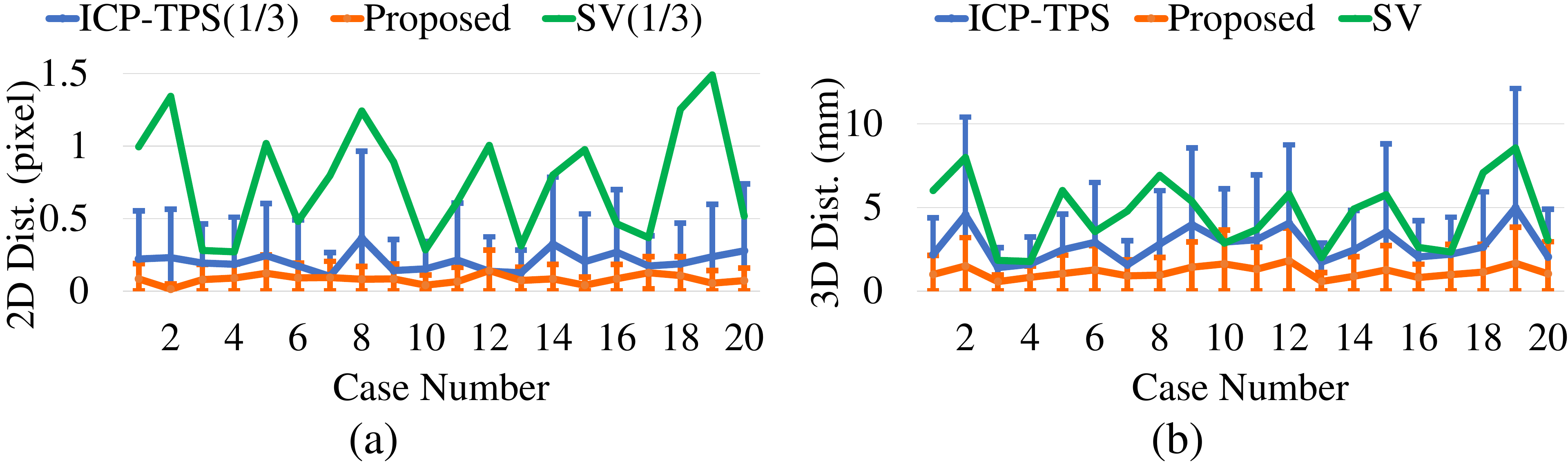}}}
    \caption{The comparison of the performance in 20 cases of experiment based on phantom data between the proposed method, ICP-TPS and the shape variation (SV), with mean$\pm$std of 2D(a)/3D(b) distance errors in 2D(a)/3D(b) environment.}
    \label{fig:pha_result}
\end{figure}

\begin{figure}[th]
    \centering
    \framebox{\parbox{3.0in}{\includegraphics[width = 3.0in]{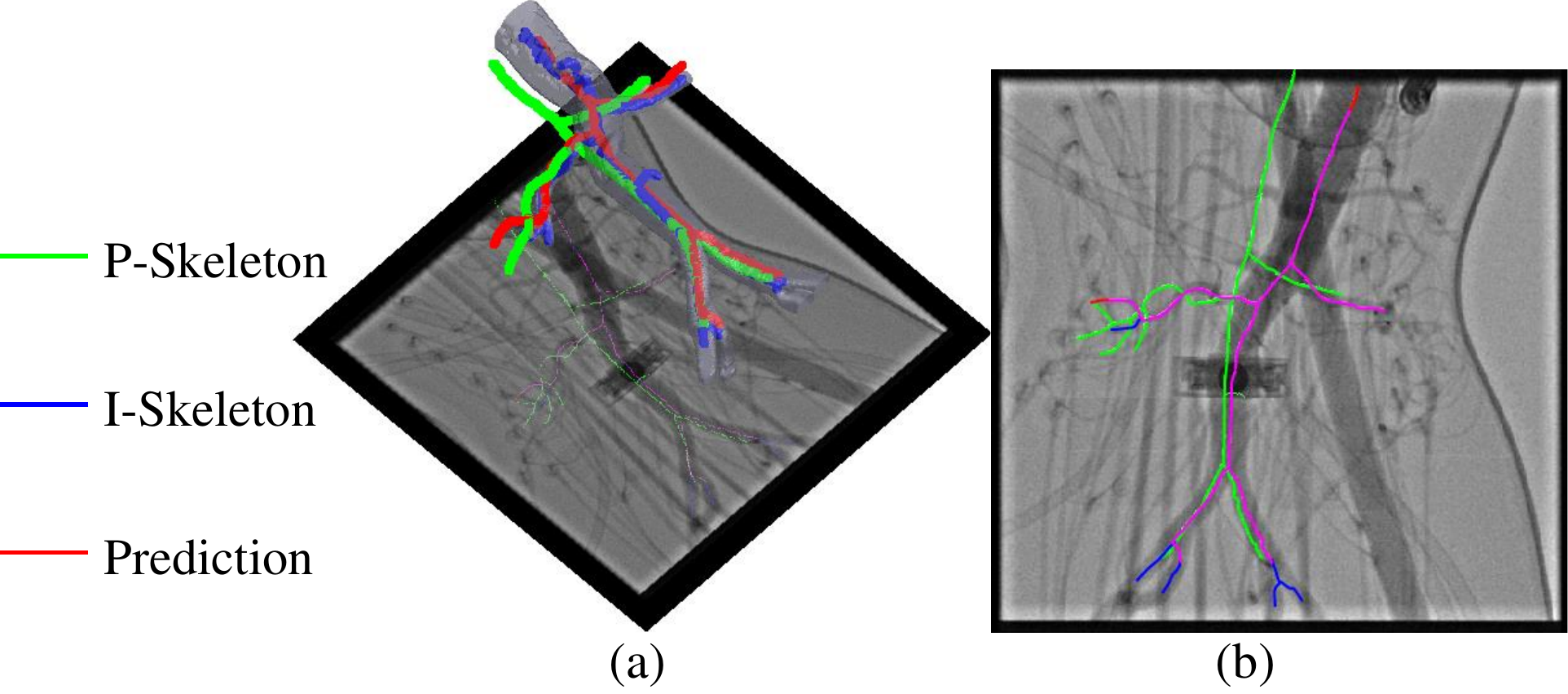}}}
    \caption{The demo of the non-rigid registration for the phantom data in 3D (a) and projected 2D (b), same abbreviations \cref{fig:sim_example} are used.}
    \label{fig:pha_example}
\end{figure}

One pre-operative 3D phantom AAA skeleton was paired with the other four intra-operative 2D phatom AAA skeletons, presenting 20 cases. The \cref{fig:pha_result}a and \cref{fig:pha_result}b show the performance of proposed method in 2D/3D environment of 20 cases of experiments based on phantom data, presented with the mean$\pm$std 2D/3D distance errors. The results show that higher accuracy, with lower 2D/3D distance errors, had been achieved by the proposed method than ICP-TPS in most cases.
The \cref{fig:pha_example} illustrated a demo of the non-rigid registration by the proposed method in experiment based on phantom data, where proper registration was achieved with predicted deformation.
\subsection{Patient Data Experiment}\label{sec:result_patient}
\begin{figure}[th]
\centering
\framebox{\parbox{3.3in}{\includegraphics[width = 3.3in]{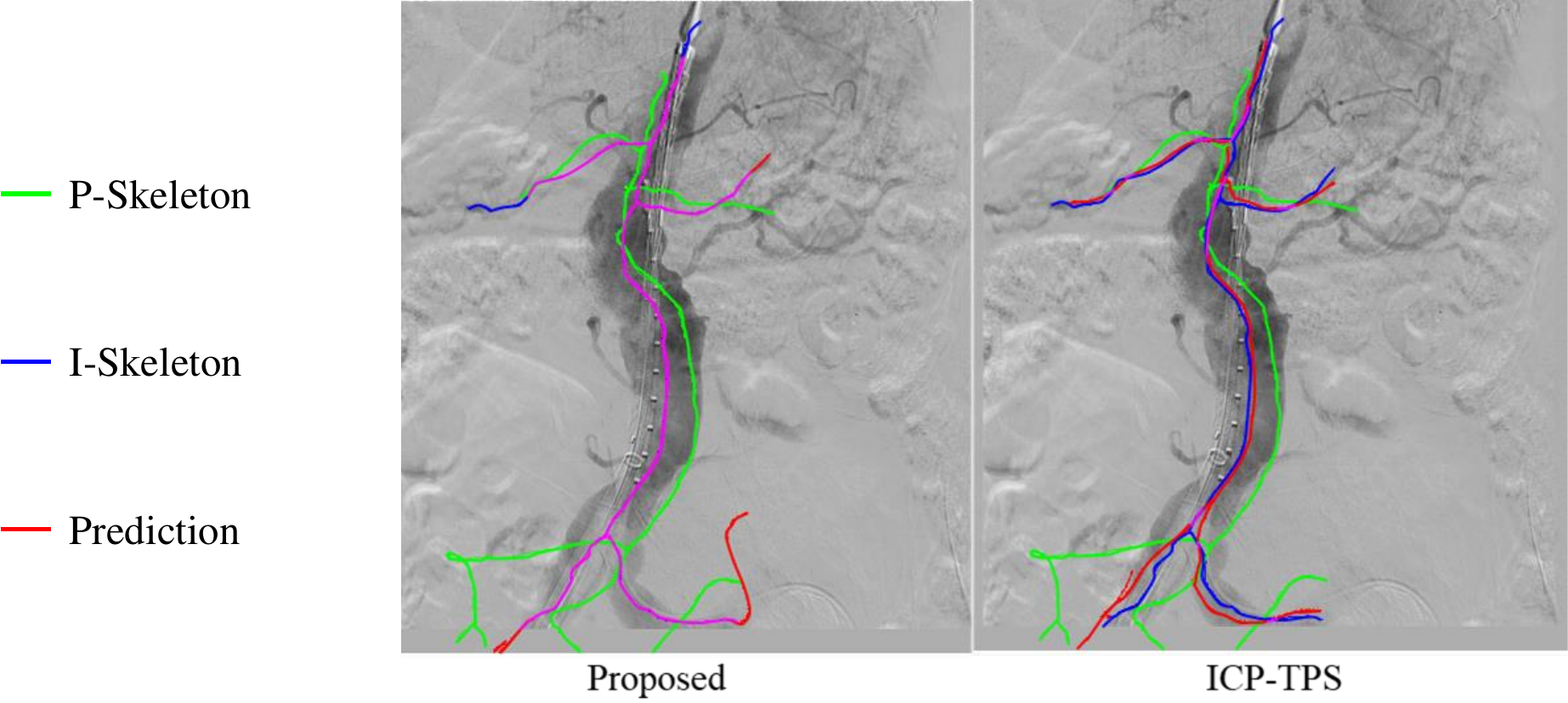}}}
\caption{The demo of the registration by the proposed method and ICP-TPS in the 2D projection plane, same abbreviations \cref{fig:sim_example} are used.}
\label{fig:pat_example}
\end{figure}

The pre-operative 3D AAA skeletons and corresponding intra-operative 2D skeletons of four patients were used for the validation. The performance of the proposed method in 2D environment was validated by the experiment on four groups of patient data, which was evaluated only by mean and std of 2D distance errors shown in \cref{tab:pat_result}, since the ground truth in 3D intra-operative cases was impossible to acquire. The result shows that higher accuracy of registration, with lower 2D distance error and std, by the proposed method have been achieved in these experiment on patient data than ICP-TPS, the baseline.
\begin{table}[ht]
	\centering
	\caption{2D distance errors (2D dist.) achieved by the ICP-TPS and the proposed method and Shape Variation (SV) for four patients.}
    \label{tab:pat_result}
    \begin{tabular}{cccc}
    \hline
        \multicolumn{2}{c}{}	& \multicolumn{2}{c}{2D dist. (pixel)}	\\\cline{3-4}
        Patient                 & Method 		& avg.  & std.  \\ \hline\hline
        \multirow{3}{*}{1}	    &SV		        &9.4509	&- 	    \\\cline{2-4}
                       		    & ICP-TPS		&0.6357	&1.1966	\\\cline{2-4}
                       		    & Proposed   	&0.1309	&0.1482 \\ \hline
        \multirow{3}{*}{2}	    &SV		        &11.1678&- 	    \\\cline{2-4}
                       		    & ICP-TPS		&0.6200	&2.1370	\\\cline{2-4}
                       		    & Proposed   	&0.1268	&0.1554 \\ \hline
        \multirow{3}{*}{3}	    &SV		        &4.6217	&- 	    \\\cline{2-4}
                       		    & ICP-TPS		&0.5364	&1.2741	\\\cline{2-4}
                       		    & Proposed   	&0.1127	&0.1488 \\ \hline
        \multirow{3}{*}{4}	    &SV		        &8.9363	&- 	    \\\cline{2-4}
                       		    & ICP-TPS		&1.0325	&3.0812	\\\cline{2-4}
                       		    & Proposed   	&0.1185	&0.1455 \\ \hline
    \end{tabular}
\end{table}
\cref{fig:pat_example} illustrates a demo of the non-rigid registration by the proposed method and ICP-TPS in the experiment based on patient data, where proper registration, with higher overlap in 2D image, was achieved with predicted deformation.
Tab.~\ref{tab:reg_result} summarizes the performance of the ICP-TPS and proposed method in experiment on the simulation (Sec.~\ref{sec:result_sim}), phantom (Sec.~\ref{sec:result_phan}) and patient data (Sec.~\ref{sec:result_patient}) by comparing 2D/3D distance errors. The results conclude that the proposed method outperformed ICP-TPS with lower 2D/3D distance error and std.
\begin{table}[th]
	\centering
	\caption{The overall performance of registration by the ICP-TPS and the proposed method, and Shap Variation (SV) in experiment on the simulation, phantom and patient data, compared by the mean 2D and 3D distance errors (2D and 3D dist.) and std.}
    \label{tab:reg_result}
    \begin{tabular}{cccc}
        \hline
        Dataset             & Method 	    & {2D dist. (pixel)}& {3D dist. (mm)}\\
        \hline\hline
        \multirow{3}{*}{Simulation}	&SV		&2.3119	            &4.6431 \\\cline{2-4}
                       		& ICP-TPS		&0.6247	            &2.7600	\\\cline{2-4}
                       		& Proposed   	&0.0817	            &1.1323 \\ \hline
        \multirow{3}{*}{Phantom}	&SV		&2.1628             &5.3186 \\\cline{2-4}
                       		& ICP-TPS		&0.9188	            &3.2949	\\\cline{2-4}
                       		& Proposed   	&0.1196	            &2.0180 \\ \hline
        \multirow{3}{*}{Patient}	&SV		&4.6217	            &- 	    \\\cline{2-4}
                       		& ICP-TPS		&0.7061	            &-	    \\\cline{2-4}
                       		& Proposed   	&0.1222	            &-      \\ \hline
    \end{tabular}
\end{table}
% \textcolor{qb}{The 2D skeletonization by the proposed method took about 0.9s, while the non-rigid registration with 700--900 graph nodes costED 0.8s with the proposed method but took over 10 mins with the ICP-TPS, where all program were run on ${MATLAB}^{\textregistered}$ with a 3.6GHz$\times$8 CPU (${Intel}^{\textregistered}$ Core.}

About 0.9s was spent on 2D skeletonization , as well as the non-rigid registration with 700--900 graph nodes costed about 0.8s for the proposed method and over 10 mins for the ICP-TPS, based on ${MATLAB}^{\textregistered}$ with a 3.6GHz$\times$8 CPU (${Intel}^{\textregistered}$ Core).

\section{Discussion}
\label{sec:discussion}
A 3D robotic path could be constructed by the proposed framework in this paper from pre-operative 3D CT scans and intra-operative 2D fluoroscopic images of AAA. Among the experiments in simulation, phantom and patient data in Sec.\ref{sec:result}, the proposed method has achieved better performance than ICP-TPS in non-rigid registration, with lower 2D/3D distance error, robustness in frontal translation and rotation, and better time-efficiency. Since fluoroscopic image required $0.2 \sim 0.5$ seconds per image to capture during FEVAR, the proposed method implemented with a more efficient programming language is potential to realize real-time 3D navigation within 1s time consumption.

The proposed graph matching is based on topology, implying that it is sensitive to topology variance. The five specified main branches are required to be the longest ones in a skeleton during the registration. and other branches could be done semi-automatically with our open-source code. The proposed method might also be sensitive to transverse and sagittal transformations, as branch node assigning based on the projected path length of 3D branches. In practical application, this should not be a problem, as the C-arm is usually fixed at a frontal scan to save space for surgeons.

The deep learning method for pre-operative 3D AAA segmentation is only briefly introduced in Sec.\ref{sec:extraction}, as we could only segment the main aortic aneurysm with reasonable results. Manual corrections might be required for aortic branches including the renal arteries and illiac branches. In the future, we will focus on a complete segmentation of the aortic aneurysm with deep convolutional neural networks.

\section{CONCLUSIONS}
\label{sec:conclusion}
A real-time framework for 3D robotic path planning is proposed to instantiate the 3D intra-operative AAA skeleton from a single 2D AAA fluoroscopic image. Pre-operative 3D AAA skeleton is deformed to match the intra-operative 2D AAA skeleton with graph matching and smooth/length maintenance. Furthermore, deep learning is used to facilitate the framework automation. Comparable accuracy, robustness and time efficiency are achieved, however the proposed method could suffer from influences of topological variance and transformation in the saggital and transverse plane.
	
\section*{ACKNOWLEDGMENT}
This work is supported by Engineering and Physical Sciences Research Council (EPSRC) project grant EP/L020688/1 and EP/N024877/1. We gratefully acknowledge the support of NVIDIA Corporation with the donation of the Titan Xp GPU used for this research.

\bibliographystyle{IEEEtran}
\bibliography{reference}

\end{document}